# Computational analyses of the topics, sentiments, literariness, creativity and beauty of texts in a large Corpus of English Literature

Arthur M. Jacobs[1,2] & Annette Kinder[3]

Author Note

1 Experimental and Neurocognitive Psychology, Freie Universität Berlin, Germany

2 Center for Cognitive Neuroscience (CCNB), Freie Universität Berlin, Germany

3 Learning Psychology, Freie Universität Berlin, Germany

Correspondence: Arthur M. Jacobs

Department of Experimental and Neurocognitive Psychology, Freie Universität Berlin, Habelschwerdter Allee 45 , D-14195 Berlin, Germany.

Email: ajacobs@zedat.fu-berlin.de

**Abstract**

The *Gutenberg Literary English Corpus (GLEC*, Jacobs, 2018a*)* provides a rich source of textual data for research in digital humanities, computational linguistics or neurocognitive poetics. However, so far only a small subcorpus, the *Gutenberg English Poetry Corpus,* has been submitted to quantitative text analyses providing testable predictions for scientific studies of literature. Rather than focussing on only a single type of literature, in this study we address differences among the different literature categories in GLEC, as well as differences between authors. We report the results of three studies providing i) topic and sentiment analyses for six text categories of GLEC (i.e., children and youth, essays, novels, plays, poems, stories) and its >100 authors, ii) novel measures of semantic complexity as indices of the literariness, creativity and book beauty of the works in GLEC (e.g., Jane Austen's six novels), and iii) two experiments on text classification and authorship recognition using novel features of semantic complexity. The results of study 1 revealed essays and plays as the 'happiest' and 'unhappiest' categories of GLEC, respectively. The 'happiest' authors for novels and poetry were Walt Whitman and John Keats, the 'unhappiest' in both categories Rudyard Kipling, while William Shakespeare wrote the unhappiest and most fearful *plays* in this sample. Study 2 showed first that the phenomenon *vocabulary decay* is general to all six categories of GLEC, while prolific writing by itself (e.g., >20 works in GLEC) does not increase its likelihood, time pressure (e.g., Charles Dickens vs. Jane Austen) being the crucial factor (Elliott, 2016). Second, the data on two novel measures estimating a text's *literariness*, intra-textual variance and stepwise distance (van Cranenburgh et al., 2019) revealed that plays are the most literary texts in GLEC, followed by poems and novels. Computation of a novel index of text *creativity* (Gray et al., 2016) revealed poems and plays as the most creative categories with the 'most creative' authors all being poets (Milton, Pope, Keats, Byron, or Wordsworth). Third, for the first time we computed a novel index of perceived beauty of verbal art (Kintsch, 2012) for the works in GLEC and predict that 'Emma' is the theoretically most beautiful of Austen's novels. Finally, study 3 demonstrated that these novel measures of semantic complexity are important features for text classification and authorship recognition with overall predictive accuracies in the range of .75 to .97. Our data pave the way for future computational and empirical studies of literature or experiments in reading psychology and offer multiple baselines and benchmarks for analysing and validating other book corpora.

# 1 Introduction

Text corpora are an indispensable tool for scientific studies in many fields such as natural language processing (NLP), digital humanities and literary studies, psycholinguistics or neurocognitive poetics (Kucera & Francis, 1967; Neidorf et al., 2019; Samothrakis & Fasli, 2015; Jacobs, 2015, 2018a,b). They offer the textual basis for tackling numerous novel research questions, such as:

- why *Molière* most likely did write his plays (Cafiero & Camps, 2019)
- what the dominant shapes of the emotional arcs of stories are (Reagan et al., 2016)
- how happy people were during previous centuries (Hills et al., 2019)
- whether the *Pollyanna effect*[1] can be found in German and English children and youth literature (Jacobs et al., 2020)
- how literariness ratings can be predicted (van Cranenburgh et al., 2019)
- whether different dramatic subgenres in French dramas of the classical age have distinctive dominant topics (Schöch, 2017)
- which topics dominate the 154 Shakespeare sonnets (Jacobs et al., 2017)
- or which text features mainly control readers eye movements during reading such sonnets (Xue et al., 2019, 2020).

Especially empirical studies investigating not only texts themselves –as in NLP or stylometrics– but behavioral or neuronal reader responses to texts, like in neuroimaging studies on reading passages from the *Harry Potter* series (Hsu et al., 2015; Wehbe et al., 2014), can benefit from results on quantitatively analysed text corpora, because they provide predictions based on the number and type of salient features systematically discriminating one type of text from another, or inducing one dominant mood in a given set of poems (e.g., Lüdtke et al., 2014).

# 2 Corpus: GLEC

The *GLEC* (Jacobs, 2018a) is such a tool offering plenty of novel possibilities for stylometric or empirical studies of literature with a high potential for providing benchmarks, for example on authorship recognition performance. It contains ~900 novels, 500 short stories, 350 tales and stories for children, 330 poetry collections, poems and ballads, 130 plays, as well as 550 pieces of non-fiction, e.g. articles, essays, lectures, letters, speeches or (auto-)biographies, with ~*12 million* sentences and *250 million* words from a wide range of authors such as Austen, Byron, Coleridge, Darwin, Dickens, Einstein, Eliot, Poe, Twain, Woolf, Wilde, or Yeats. Previous work presented the results of stylometric, topic and sentiment analyses only for a subset of *GLEC*, the ~120 texts from the *Gutenberg English Poetry Corpus* motivating several behavioral experiments investigating eye movement control during the reading of Shakespeare sonnets (Xue et al., 2019, 2020). The present analyses extend this work to the whole *GLEC* providing answers to a number of basic questions that can guide future scientific studies of literature, stylometrics or reading psychology. Table 1 shows 60 example texts from *GLEC,* 10 from each of six broad categories (children and youth literature/CYL, essays/ESS, novels/NOV, plays/PLA, poems/POE and short stories/STO).

---

[1] In 1969 Boucher and Osgood presented influential evidence for the idea that "humans tend to look on (and talk about) the bright side of life" and coined this phenomenon the "Pollyanna hypothesis," i.e., a universal human tendency to use evaluatively positive words more frequently, diversely and facilely than evaluatively negative words.

**Table 1.** (60 example texts from *GLEC,* 10 from each category)

| CHILDREN AND YOUTH LITERATURE | ESSAYS |
|---|---|
| Andrew Lang.Tales of Troy and Greece | Bertrand Russell.The Analysis of Mind |
| Baronness Orczy.The Scarlet Pimpernel | Charles Darwin.On the Origin of Species 6th Edition |
| Beatrix Potter.The Tale Of Peter Rabbit | George Eliot.The Essays of George Eliot1 |
| Edward Stratemeyer.The Rover Boys in the Land of Luck | John Locke.An Essay Concerning Humane Understanding |
| Jacob Abbott.Cleopatra | John Stuart Mill.A System Of Logic |
| James Matthew Barrie.Peter Pan | Lewis Carroll.Symbolic Logic1 |
| Louisa May Alcott.Rose in Bloom | Michael Faraday.Experimental Researches in Electricity |
| Lyman Frank Baum.The Wonderful Wizard of Oz | Sir Isaac Newton.Opticks |
| R M Ballantyne.Away in the Wilderness | Sir Winston Churchill.Liberalism and the Social Problem |
| Thornton Waldo Burgess.Mrs. Peter Rabbit | William Butler Yeats.Discoveries |
| **NOVELS** | **PLAYS** |
| Bram Stoker.Dracula | George Bernard Shaw.Pygmalion |
| Charles Dickens.Oliver Twist | John Dryden.All for Love |
| D H Lawrence.Women in Love | John Galsworthy.A Bit O Love |
| Daniel Defoe.The Life and Adventures of Robinson Crusoe | Oscar Wilde.Lady Windermeres Fan |
| Edgar Rice Burroughs.Tarzan of the Apes | Richard Brinsley Sheridan.The Rivals |
| George Eliot.Middlemarch | William Butler Yeats.The Hour Glass |
| Herman Melville.Moby Dick | William Dean Howells.The Sleeping Car |
| Jane Austen.Emma | William Shakespeare. Hamlet-Prince of Denmark |
| Sir Arthur Conan Doyle.The Hound of the Baskervilles | William Shakespeare. Romeo And Juliet |
| Winston Churchill.Coniston | William Shakespeare.The Tempest |
| **POETRY** | **SHORT STORIES** |
| Alexander Pope.The Poetical Works | Charles Dickens.Holiday Romance |
| D H Lawrence.Amores | George Eliot.Brother Jacob |
| Elizabeth Barrett Browning.The Poetical Works | Henry James.The Chaperon |
| John Keats.Endymion1 | Jane Austen.Love And Friendship |
| John Milton.The Poetical Works of John Milton1 | Joseph Conrad.Youth |
| Lord Byron.Byrons Poetical Works Vol. 1 | Lucy Maud Montgomery.Short Stories 1896 to 1901 |
| P B Shelley.The Complete Poetical Works | Mark Twain.The Facts Concerning The Recent Carnival Of Crime In Connecticut |
| Samuel Taylor Coleridge.The Complete Poetical Works1 | O Henry.Rolling Stones |
| William Butler Yeats.The Wild Swans at Coole | Oscar Wilde.Shorter Prose Pieces |
| William Shakespeare__sonnet32 | Rudyard Kipling.Indian Tales |

**Corpus Preprocessing**

Our set of 2722 texts was taken from the overall ~2950 texts of GLEC: Only texts from authors that occurred at least five times in one of the six text categories were selected. This was necessary to enable the standard 5-fold cross-validation procedure for the machine learning part in part five of this paper. All texts were pre-processed using standard NLP python routines available from the open access NLTK library (Bird et al., 2009). In particular, every sentence in each text was POS-tagged and lemmatized using the treetagger routine (Schmid, 1994: https://www.cis.uni-muenchen.de/~schmid/tools/TreeTagger/) excluding stopwords from the NLTK list and only content words (nouns, verbs, adjectives and adverbs) were kept for further analyses. Using the *SentiArt* tool (Jacobs, 2017, 2019: (https://github.com/matinho13/SentiArt) each word in each sentence of each text was quantified in terms of a number of features such as word length, log frequency of occurrence in *GLEC*, emotional valence etc., and feature values were then aggregated across sentences and texts.

**3 Present study: Three Parts**

The first part of this paper presents two exploratory descriptive studies examining the *key topics* and the *global sentiment* of the six subcorpora of GLEC and its 112 authors. The 2$^{nd}$ part explores recent and novel measures of semantic complexity that can serve as indices of the potential *literariness, creativity,* and *beauty* of the GLEC texts. The results of these analyses reveal important differences in these global characteristics that can serve various purposes of scientific studies of literature, e.g. investigating emotional or aesthetic reader responses to given texts from a given genre or author. The final part presents two experiments on *text classifcation* and *author recognition* based on a novel set of five semantic complexity features which demonstrate their usefulness for such standard NLP tasks.

**3.1 Study 1. Topic and Sentiment Analyses**

The aim of this initial study is to identify generic topics in each of the six subcorpora or text categories of GLEC that can be discovered via a standard topic modeling tool (LDA Mallet; Shawn et al., 2012). Typically topic and sentiment analysis are separate issues in NLP. Here we show that they can be usefully combined to determine the dominant global sentiment of both subcorpora and authors.

## 3.1 Subcorpora (text categories)

**Topics**

Topic analyses were run using the gensim library (Rehurek & Sojka, 2010) with the following hyperarameter set adapted from Schöch (2017): number of topics = 50, number of words/topic = 50, number of iterations = 2500, topic threshold = .001. Based on the results of pilot studies estimating the coherence of the topics (Röder et al., 2015), only nouns (no proper names) and verbs with a minimum frequency of 100 were included.

**Figure 1 here**

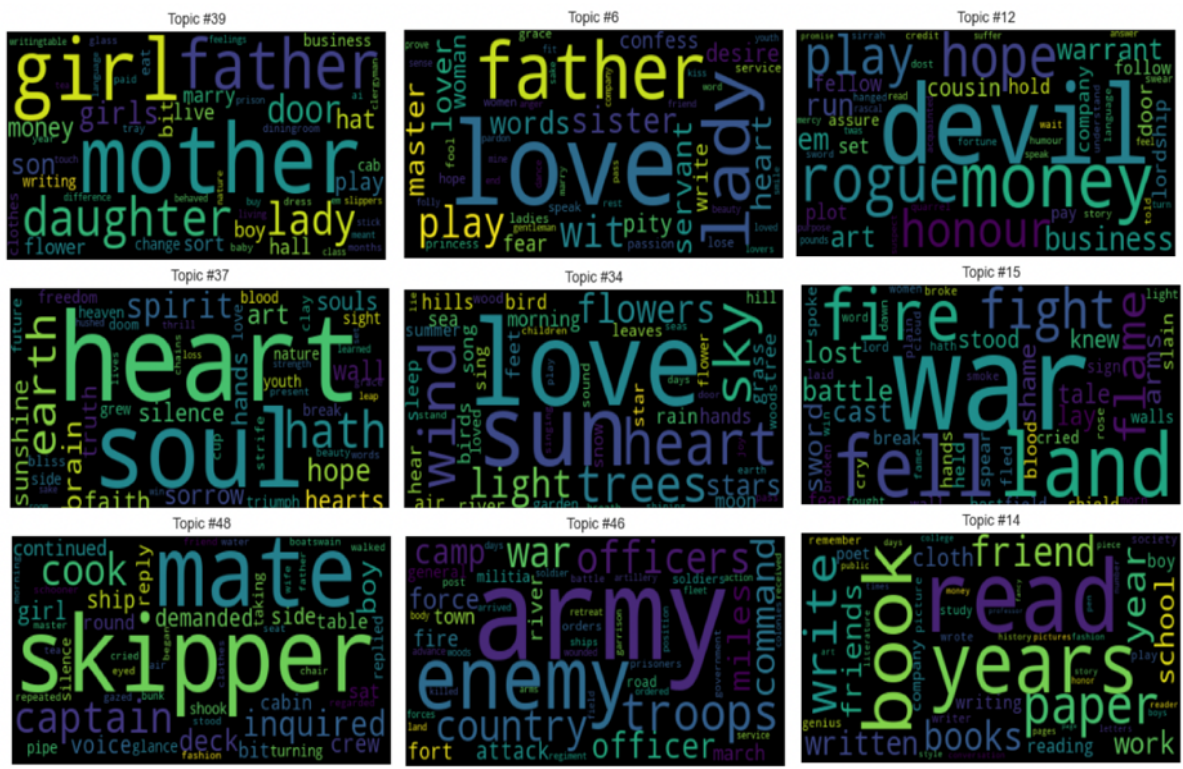

Figure 1 shows the top topics for each of the six text categories (alphabetical order from top to bottom: CYL, ESS, NOV, PLA, POE, STO) in the practical form of word clouds. They illustrate the 50 highest-ranked words of the topic and thus nicely bring out their internal structure (cf. Schöch, 2017). It can be seen that children and youth literature (CYL, top row) is dominated by marine adventure, family and circus topics. Essays (ESS, 2$^{nd}$ row) highlight scientific topics of flora, energy and knowledge. In novels (NOV, 3$^{rd}$ row) topics of marine adventure, family and army dominate. Plays (PLA, 4$^{th}$ row) turn around family, love and money topics, whereas spirituality, love and war are major poetry topics (POE, 5$^{th}$ row). Finally, the already popular marine adventure and army topics are also dominant in stories (STO, bottom row) together with a reading topic. These key topics of the 19$^{th}$ century literature presenting the core of GLEC may not generalise to contemporary literature, but they provide a solid basis for comparison with related corpora of different language, domain, or period (e.g., the German childLex corpus, Schroeder et al., 2015, or the French Poetry Corpus; Fechino et al., 2021).

**Global Sentiment: Which is the 'happiest' category and who the most 'hopeful' author?**
Recent advances in computational sentiment analyses have shown that genre can influence the emotional valence and emotional plot development of texts (e.g., Kim et al., 2017). These latter authors found computational evidence for the idea that the basic emotion *fear* is most uniform in all five but one genres investigated by them within a text corpus of ~2020 stories sampled from the Gutenberg project (adventure, humor, mystery, romance, science fiction). The exception were mystery stories, where *anger* was the most stable basic emotion. Here we applied an advanced sentiment analysis based on *SentiArt* (Jacobs, 2017, 2019) to the six much broader categories of GLEC. Such a computational corpus-based grounding of categories, genres or subgenres –using either topic or sentiment analysis– is a useful complement to the more theoretically grounded concepts of genre in theoretical literary studies (e.g., Herrmann et al., 2021; Kim et al., 2017; Schöch, 2017).

*SentiArt* is an unsupervised method based on a distributed semantic model (DSM) generated by a neural net via *gensim* (Rehurek & Sojka, 2010) trained on the entire GLEC. It simply computes the semantic relatedness of every word in a text with a set of affectively negative and positive labels (see Appendix in Jacobs, 2018a,b) and was empirically cross-validated in several previous studies (Jacobs, 2017, 2019; Jacobs & Kinder, 2018, 2019a,b; Jacobs et al., 2020). *SentiArt* quantifies the affective-aesthetic potential (AAP) and six discrete emotion values (e.g., joy, fear) for each word, sentence or paragraph of a text. For the present topic-based analyses these values were computed sentence-wise based on lemmata for nouns and verbs for each of the texts in a category.

**Figure 2 here**

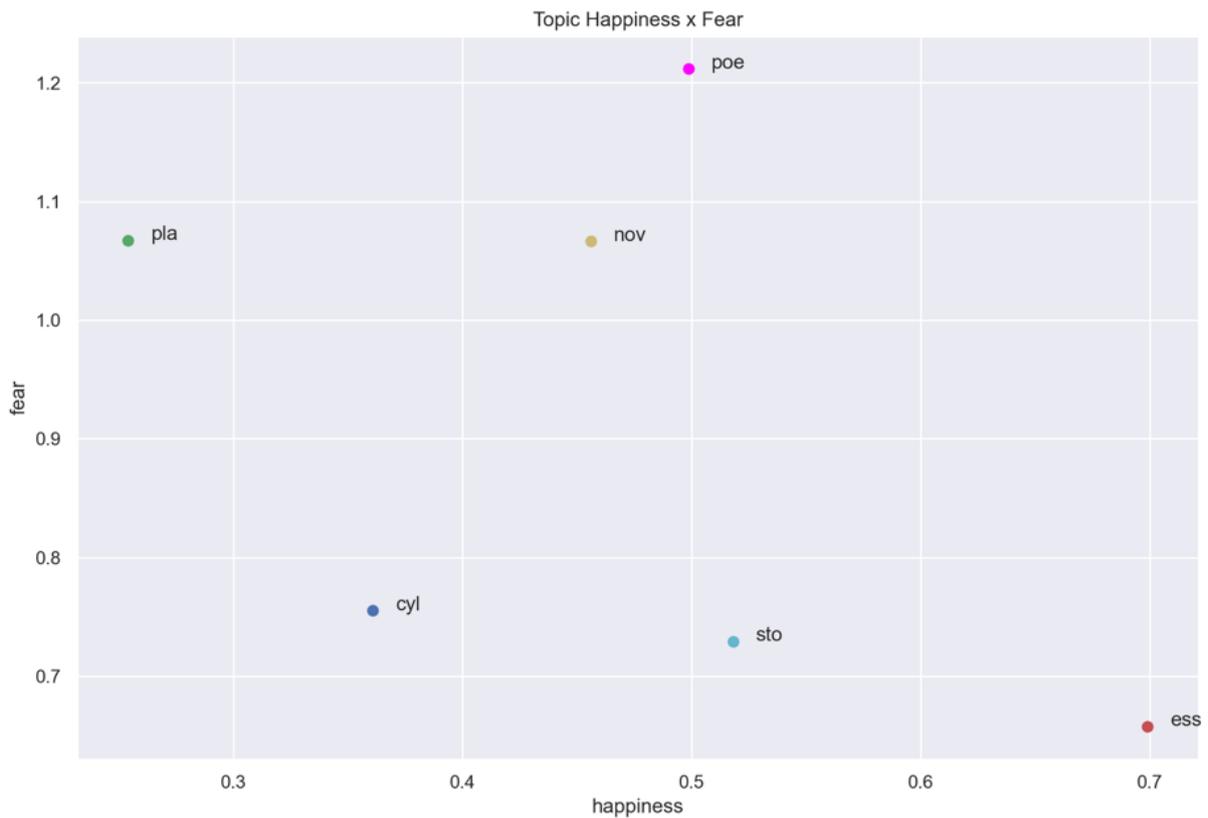

Figure 2 summarizes the results for the six text categories regarding the two basic discrete emotions 'happiness' and 'fear'. The 'happiness' scores were estimated by the mean AAP value of the 50 dominant topics represented by 50 words each. The data suggest that essays provide the 'happiest' (or most positive) topics with a mean value of .69, due to the fact that essay topics practically lack words with negative emotional connotations compared to the other five categories. They also appear to be the least fearful, i.e. most hopeful texts (hope being the emotional opposite of fear according to belief-desire theory, Reisenzein, 2009) with a mean of .65. Plays represent the other extreme having the relatively 'unhappiest' topics with a minimum value of .25, indicating lots of negative words, and the 2[nd] highest value for fear (1.07). Poems on the other hand appear to provide an interesting emotional mix between happiness and fear, featuring the most fearful (1.2) and 3[rd] happiest topics (.5). Stories are the 2[nd] happiest (.52) and 'hopeful' category (.73), while novels present a mix of happiness (.45) and fear (1.06) similar to poetry. Finally, CYL presents the 2[nd] 'unhappiest' (.36) and 3[rd] hopeful (.75) category. Statistically significant differences exist between all categories (at least p<.01), except between stories and poetry. Poetry, novels and stories appear to strike a balance between happiness and fear, while the other three categories take more extreme positions in this 2d graph.

**3.2 Authors**

We can also ask who the most 'hopeful' authors are in certain text categories and which topics they favor. Figure 3 gives a summary of the results showing the top 'happiest' and 'unhappiest' authors for each text category. Thus, Louisa May Alcott appears as the 'happiness' leader of the field for *CYL* books –of which she authored 15 in GLEC–, but also the most fearful. In contrast, Beatrix Potter's overall happiness and fear scores are both the lowest in our sample. For *essays* (not all works are annotated for better visibility), the respective heads and tails on the happiness dimension (with about equal values for fear) are Oscar Wilde and Daniel Defoe, who's 'History of the plague in London' may have contributed to his position in this 2d graph. Charles Kingsley and John Locke apparently contributed the most fearful essays, while Michael Faraday's and Charles Darwin's seem relatively hopeful.

Benjamin Disraeli and Nathanel Hawthorne, both in the lower range of the fear dimension, appear to have written the 'happiest' *novels* in GLEC in contrast to John Bunyan's (also the most fearful writer) and Daniel Defoe's overall unhappiest pieces. Interestingly, Jane Austen appears as second most fearful author in this sample. The happiest *stories* in GLEC came from Walt Whitman and Harriet Elizabeth Beecher, while Rudyard Kipling and Jack London's narratives have very unhappy contents. Again, Jane Austen with her 'Love and Friendship and Other Early Works' takes the lead on the fearful dimension topping George Eliot as 2$^{nd}$, while O'Henry and Philip Kindred Dick contributed the most hopeful stories.

Finally, William Shakespeare wrote the unhappiest and most fearful *plays* in contrast to James Matthew Barrie's happiest and hopeful ones. Shakespeare's lead on fearful texts also applies to the GLEC *poetry* sample. John Keats wrote the happiest, Rudyard Kipling the unhappiest poems and Walt Whitman the most hopeful ones.

**Figure 3 here**

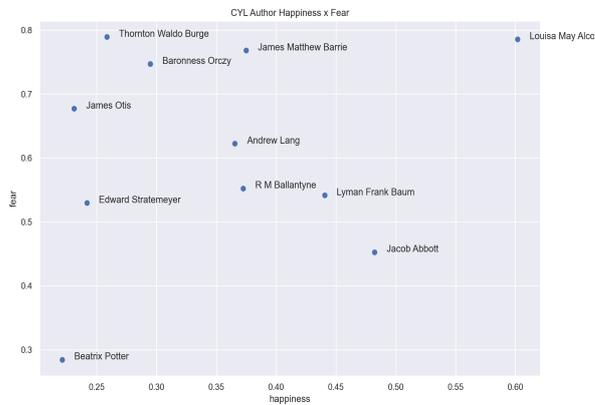
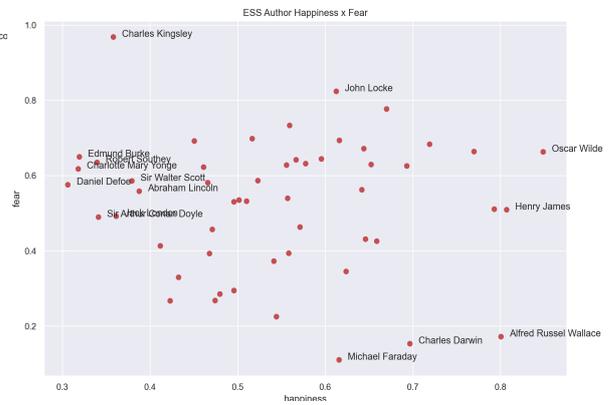
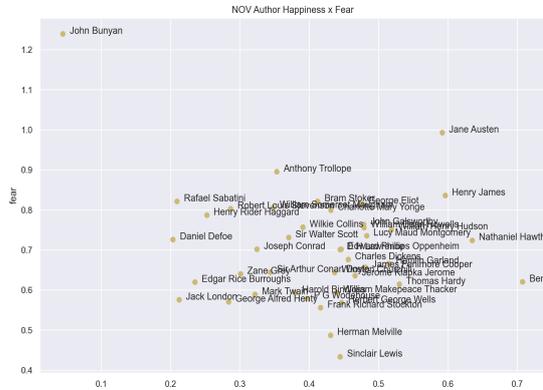
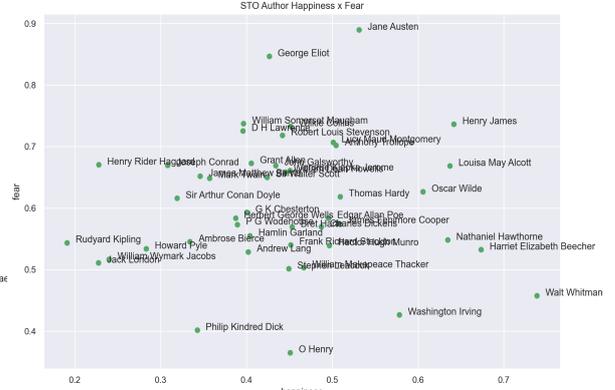
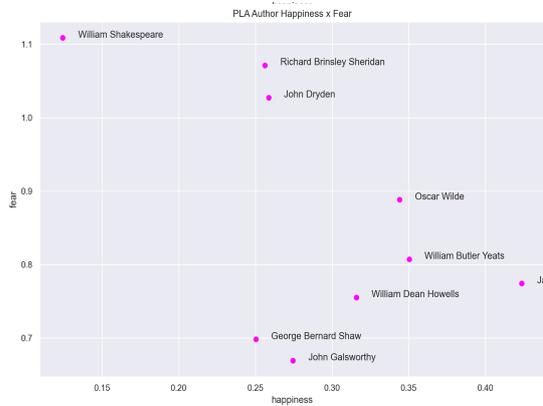
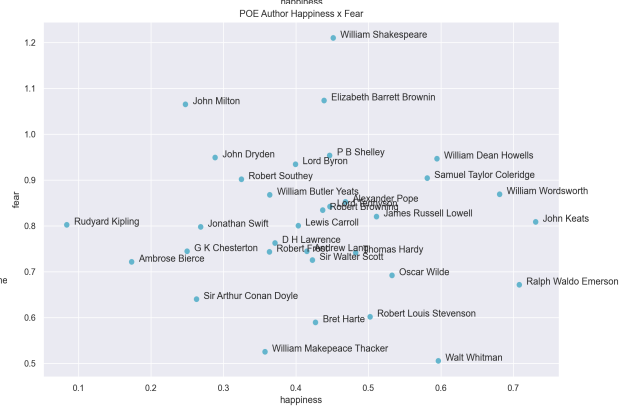

A few contrasting examples from author topic analyses matching the authors' sentiment analysis profiles are shown in Figure 4. Thus Alcott's (happiest CYL author, top row, left) favorite topic is about family affairs, while Bunyan (unhappiest novel author, top row right) prefers sin and death. For James Matthew Barrie's 'happiest' plays (middle row, left) a romantic topic featuring ladies, gardens and hope is characteristic, while Daniel Defoe's essays (middle row, right) may make readers relatively unhappy with topics like 'plague and bills'. Finally, while Rudyard Kipling's stories (bottom row, left) often turn around hunting and killing, John Keats (bottom row, right) delights his readers of poetry with eternal topics of nature like 'heaven, earth and sea'.

Of course, the data shown in the above figures reflect only coarse central tendencies observed in GLEC but do not characterise all works of a given author whose repertoire may feature many other topics expressing a whole variety of sentiments not captured in the data shown above. Still, these data can be used in future empirical studies of literature or neurocognitive poetics (Jacobs, 2015; Willems & Jacobs, 2016), e.g. for selecting textual stimuli supposed to elicit certain sentiments or reading behaviors, like the collection of Shakespeare sonnets in Xue et al.'s (2019, 2020) eye tracking studies. Moreover, while the present analyses are purely exploratory in nature, future studies may use such subcorpus- and author-based sentiment and topic analyses to predict book and author popularity or facilitate genre and authorship recognition tasks (see part 5 below).

So far our text analyses did not directly speak to the literary quality of the books in a subcorpus or by a certain author. Still, one might suspect that 'literariness' (Appel and Hanauer, 2021; Salgaro, 2018) is not equally distributed across the six text categories or the ~200 authors examined here. The next section examines this hypothesis.

**Figure 4 here**

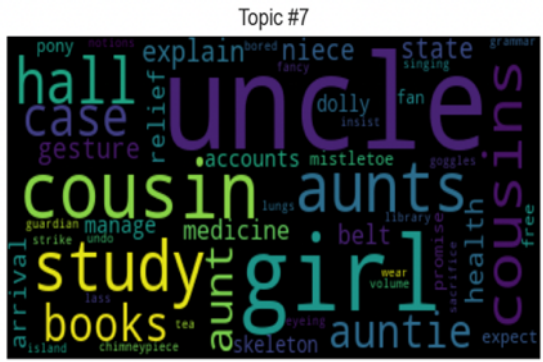
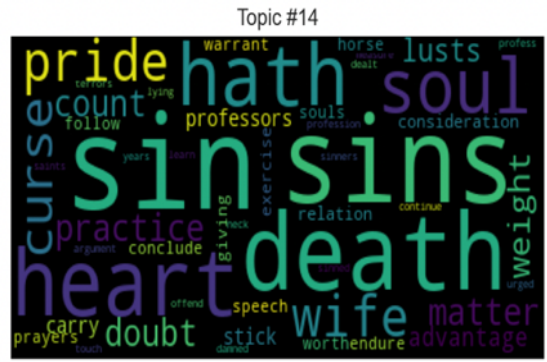
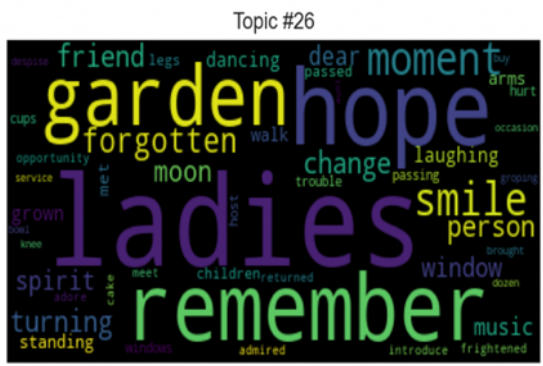
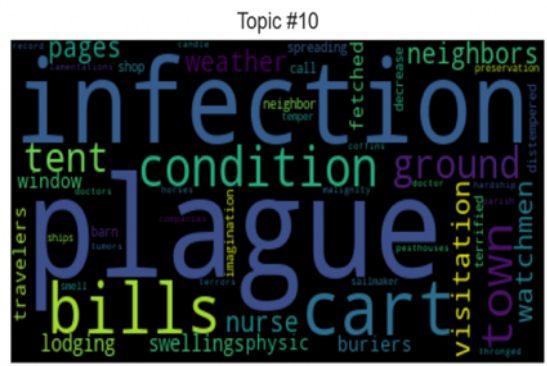
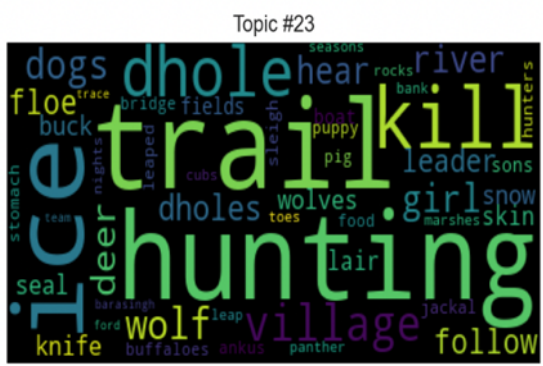
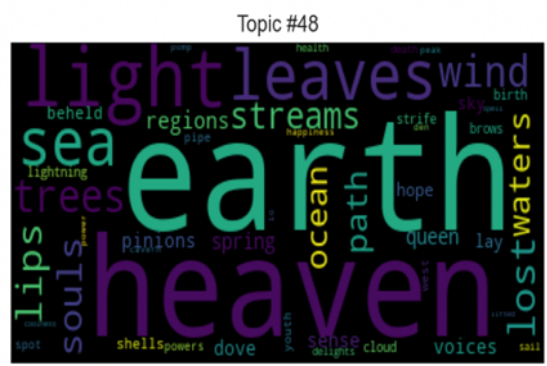

**4 Study 2: Semantic Complexity Measures**

The literature on computational stylistics offers a wealth of indices of linguistic or semantic complexity as potential indices of literary quality ('literariness','poeticity') or aesthetic success (e.g., Simonton, 1990). A classical one is *lexical diversity* which is usually measured by the *type-token ratio/ TTR* – and/or the *adjective-verb quotient/AVQ* (Jacobs, 2018b). More recently, *vocabulary decay* –the decrease in the deployment of unique words over the span of texts, as measured by Shannon's entropy index H– has been proposed as an additional index of lexical diversity (Elliott, 2016).

More complex measures use vector space models (Landauer and Dumais, 1997; Mikolov et al., 2013): Interpreting document vectors of texts as coordinates in vector space, one can apply various distance metrics to provide an operationalization of contextual (dis)similarity among these document vectors. Recent research offers derivatives of these measures to estimate the semantic complexity of texts, such as an index of their *literariness* (van Cranenburgh et al., 2019) or an index of their *creativity* (Gray et al., 2019). In study 2 we apply these measures to GLEC and supplement them by a novel measure based on a proposal by Kintsch (2012) for indexing book *beauty*. The different measures are summarized in Table 2,

**Table 2. Measures of semantic complexity with references and formulas.**

| Measure | Author(s) | Indexing | Formula(s) | Equ. # |
|---|---|---|---|---|
| **Vocabulary decay,** as estimated by Shannon entropy (H) | Elliott (2016) | Literariness | $$H(X) = -\sum_{i=0}^{n} P(x_i) \log P(x_i)$$ | (1) |
| **Intra-textual Variance** | van Cranenburgh et al. (2019) | Literariness | $$\text{variance}(T) = 1/|T| \sum_{i=0}^{i<|T|} \|\mu_T - T_i\|^2$$ where $\mu_T$ is the centroid of text T and $t_i$ is its ith chunk | (2) |
| **Stepwise Distance** | van Cranenburgh et al. (2019) | Literariness | $$\text{stepwisedist}(T) = 1/(|T|-1) \sum_{i=0}^{i<|T|-1} \|t_i - t_{i+1}\|^2$$ | (3) |
| **Forward Flow** | Gray et al. (2019) | Creativity | $$FF = \left(\sum_{i=2}^{n} \frac{\sum_{j=2}^{i-1}(D_{ij})}{i-1}\right)/(n-1)$$ where $D$ is the semantic distance between the words, and $n$ is the total number of words within a sequence | (4) |
| **Conditional Kolmogorov complexity / K(DM),** as estimated by Hellinger Bhattacharyya distance (H) | Kintsch (2012) | Beauty | $$H(P,Q) = \frac{1}{\sqrt{2}} \sqrt{\sum_{i=1}^{k}(\sqrt{p_i} - \sqrt{p_i})^2}$$ for two discrete probability distributions, $P = (p_1, \ldots, p_k)$ and $Q = (q_1, \ldots, q_k)$ | (5) |
| **Harmony** | Kintsch (2012) | Beauty | $\min_i K(D_i M_i) = \text{mean}(K(D_i M_i))$ | (6) |
| **Variety** | Kintsch (2012) | Beauty | $K(A_i M \mid A_j M) > K(A_i M, A_j M \mid DM)$ where D is a real-world object (e.g., a book) with aspects $A_1, \ldots A_i, \ldots A_k$ (e.g., chapters), M is the mental model of that object, i.e. the result of a human observer using a particular encoding procedure (M) to encode the object and its aspects, and DM is a mental representation of the data D given the model M. | (7) |

## 4.1 Entropy and vocabulary decay

As argued by Elliott (2016), the higher the entropy (H; cf. equ. 1 in Table 2) of the vocabulary of a text or of one of its part's (e.g., beginning or ending), the wider the vocabulary range, or its diversity. Elliott's main finding was that novels written under time pressure such as those of Charles Dickens display a characteristic contraction of the vocabulary (i.e., lower H values) towards their end, while novels for which the pace of writing was far from hasty, such as Jane Austen's, lack this vocabulary decay trend. Here we were interested in examining whether such systematic trends show up in the different authors and categories in GLEC.

**Authors**

As already mentioned, Jane Austen's novels showed no vocabulary decay in Elliott's (2016) entropy analysis, while Dickens'novels did –an indicator of overall rather hasty writing. Here we looked at individual authors to see whose novels may have been written under time pressure, as assumed by Elliott (2016). First we checked whether our data correspond with those of Elliott (2016) for Austen's seven and Dickens' 20 novels included in GLEC. Figure 5 (upper panel, left) shows the data which confirm Elliott's findings: In contrast to Dickens', Austen's novels show no clear trend of a vocabulary decay.

**Figure 5 here**

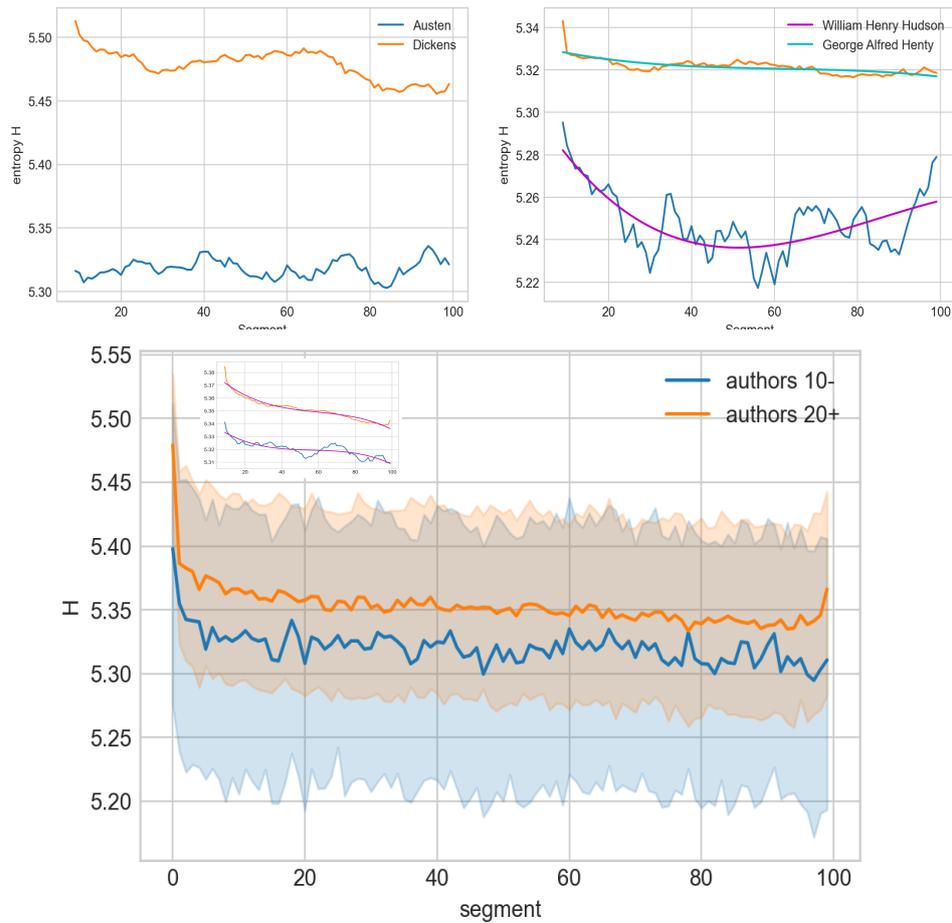

Next we tested the idea whether vocabulary decay is perhaps typical for very prolific authors ('writing maniacs'). Out of overall 42 authors of the 809 GLEC novels, 18 produced 10 or less novels, among which Jane Austen (7), George Eliot (8) or William Henry Hudson (5), while 15 produced 20 or more novels, e.g. George Alfred Henty (87!), Charlotte Mary Yonge (27), or James Fenimore Cooper (30). As can be seen in Figure 5 (lower panel) the average data do not support this idea: both curves (aut10- and aut20+) show a similar trend of vocabulary decay (data in main Figure give mean values with 95% confidence intervals, data in inlay give the smoothed average curves –window of 10– with best polynomial fits (3$^{rd}$ degree).

However, when contrasting the most prolific author (George Alfred Henty) with the least prolific one (William Henry Hudson), the idea is confirmed, as shown in Figure 5 (upper panel, right). Still, even for the most prolific author the trend is rather light. Thus, it seems safe to say –at least for the present GLEC novels– that in general prolific writing by itself does not increase the likelihood of vocabulary decay, but that one should always take a closer look at individual authors.

If time pressure enters the equation as can be assumed for the prolific author Charles Dickens (Elliott, 2016), then vocabulary decay is a likely side-effect of prolificness.

**Text Categories**

Do novels and stories differ with regard to their likelihood of vocabulary decay from poems or plays? Regarding a particularly popular type of poetry, Shakespeare sonnets, in a previous stylistics computational study we had already shown that the number of new text world referents decreases towards the end of the 154 sonnets (Jacobs et al., 2017). While the first quatrains feature ~90% new words (per sonnet) on average, this value drops to ~75% for the 2$^{nd}$ quatrains and 68% for the 3$^{rd}$ quatrains reaching a plateau for couplets with ~60% (all $p$ < .0001). This discovery shed new light on Vendler's (1997) "funnel-shape" movement assumption, a lexico-semantic narrowing down movement from *quatrain 1* (e.g., wide epistemo- logical field) to *quatrain 2* (e.g., queries, contradicts, subverts position in quatrain 1) to *quatrain 3* (e.g., subtlest, most comprehensive/truthful position and solution) to the final couplet (summarizing, ironic or expansive coda – restating semantically the body of the sonnet, i.e., quatrain 1 to 3 – with a crucial tonal difference and an often a self-ironizing turn to the proverbial or idiomatic). This finding raises the question whether such a vocabulary decay is specific for sonnets or typical for the entire of poetry, and perhaps also other genres like plays or essays.

Before we turn to the issue of category vocabulary decay, it is worth mentioning that in our sonnet study we had used the *ttr* measure which should be positively related to the entropy measure applied above. Since Elliott did not examine this, we ran an analysis on our subcorpora and indeed found significant correlations between the two measures (CYL: $r^2$ = .94; ESS: $r^2$ = .58; NOV: $r^2$ = .97; PLA: $r^2$ = .93; POE: $r^2$ = .82; STO: $r^2$ = .93; all p <.001).

Regarding the issue of vocabulary decay, we computed the average entropy for each of our six text categories, as summarized in Figure 6. The (standardized, smoothed, rolling window of 10) data exhibit an overall nonlinear (cubic) trend towards lower H values from beginning to end, with only some slight variation across categories. To determine whether these trends are statistically significant we ran a simplified analysis using only 10 segments (aggregating 10 data points per segment) by computing the polynomial fits (3$^{rd}$ degree) for each category together with an analysis of variance (ANOVA). The adjusted $R^2$ values in Table 3 indicate the goodness of fit; p values smaller than 0.05 indicate that the nonlinearly decreasing trend is significant.

**Figure 6 here**

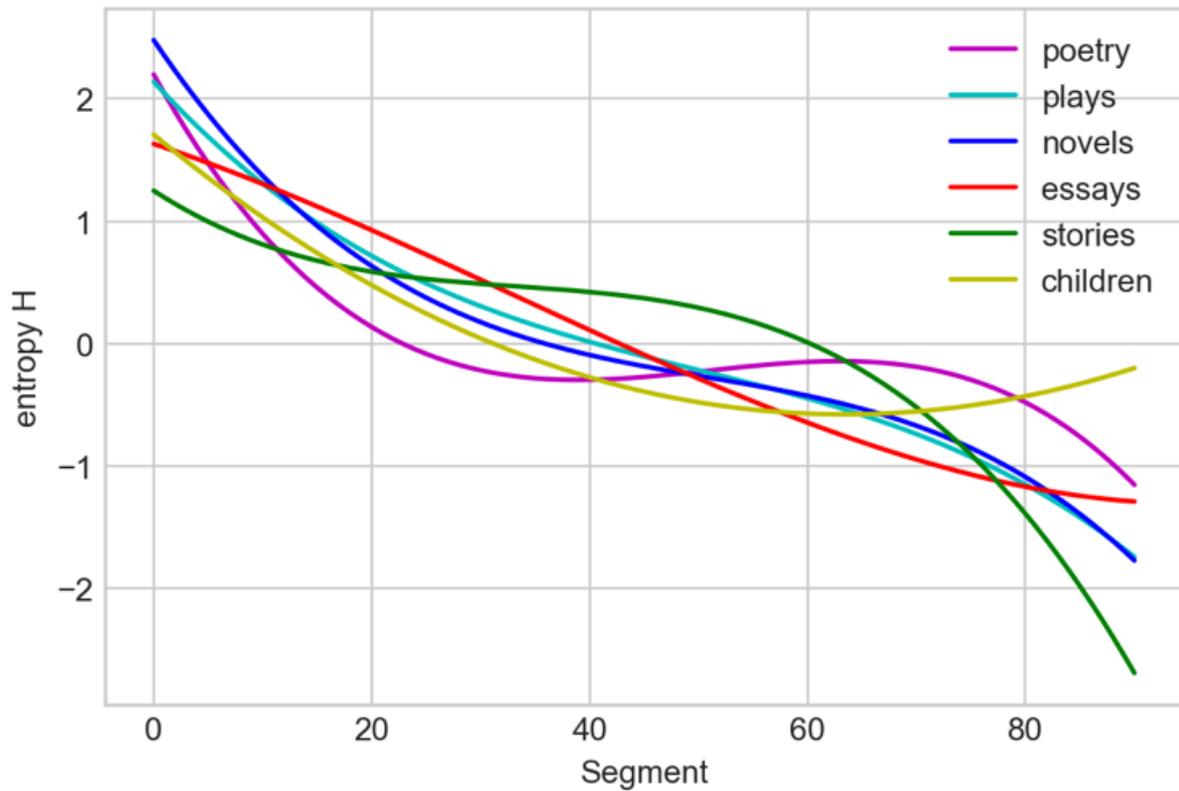

**Table 3. Goodness of fit measures (polynomial fits - 3$^{rd}$ degree) for the six text categories.**

| category | R²$_{adj}$ | F ratio | p value < x |
|---|---|---|---|
| children | .15 | 8.7 | .0003 |
| essays | .52 | 35.2 | .0001 |
| novels | .46 | 27.9 | .0001 |
| plays | .43 | 25.5 | .0001 |
| poems | .16 | 6.0 | .0009 |
| stories | .31 | 14.3 | .0001 |

To sum up this section, the phenomenon of vocabulary decay, already shown for novels written under time pressure (Elliott, 2016) and Shakespeare sonnets (Jacobs et al., 2017), appears to be a general trend of the literature represented in GLEC (Jacobs, 2018a). On average, all six text categories exhibit vocabulary decay, although individual texts by individual authors can deviate from it, e.g. Jane Austen's novels. We could find no evidence for the idea that this phenomenon is specific for prolific authors, although when contrasting extreme pairs of authors such as Dickens vs. Austen, or Henty vs. Hudson, only the more prolific ones showed vocabulary decay. Future studies could examine the phenomenon more closely by looking at finer grained genres like adventure, science fiction, or mystery novels or compare novels of the same author written under different conditions of time pressure to shed more light on its possible causes.

**4.2 Intra-textual Variance and Stepwise Distance**

Following the approach of van Cranenburgh et al. (2019) we computed intra-textual variance (ITV; cf. equ. 2 in Table 2) and stepwise distance (SWD; cf. equ. 3 in Table 2) measures for all texts in the six GLEC categories. If these two measures (which correlate with r = .85, p <.0001) are considered valid indices of literariness, then according to the data in Table 4 below plays are the most literary texts in GLEC (i.e., having the highest values of ITV and SDW), followed by poems and novels. CYL comes fourth followed by stories and essays (all p <.0001).

**Table 4. Measures of semantic complexity for the six text categories.**

| category | H* | ITV | SWD | FF | HARM | VAR |
| --- | --- | --- | --- | --- | --- | --- |
| children | 4.88 [4.9, 4.88] | 3,0698 | 4,4996 | 0,837 | 0,0882 | 0,0829 |
| essays | 4.93 [4.94, 4.91] | 2,4351 | 3,1427 | 0,839 | 0,0983 | 0,0867 |
| novels | 5.33 [5.35, 5.32] | 4,9686 | 7,3943 | 0,835 | 0,0917 | 0,0869 |
| plays | 4.58 [4.61, 4.57] | 9,7046 | 15,435 | 0,851 | - | - |
| poems | 2.37 [2.38, 2.37] | 8,349 | 14,324 | 0,865 | - | - |
| stories | 4.75 [4.76, 4.74] | 3,0866 | 3,937 | 0,837 | 0,0914 | 0,0844 |

*Global mean and [mean first chunks 1-10, mean last chunks 91-100]

Figure 7 reveals important between-author differences in the two computational indices of literariness in each of the text categories demonstrating the potential of these indices for future empirical studies of literature. As an example from poetry, in Figure 8 we zoom into the works of three authors, illustrating the stepwise distances (initial chunks only; cf. van Cranenburgh et al., 2019) and establishing the following rank order: Rudyard Kipling (mean SWD: 18.11) > Walt Whitman (16.6) > William Shakespeare (13.1). It would be a valuable task for future research to empirically test these computational predictions following exemplary studies by Lüdtke et al. (2014), Xue et al. (2019, 2020), or Gambino et al. (2020) who examined the perceived beauty or poeticity of various poetic works including Shakespeare sonnets.

It should be noted though that on the basis of their results concerning Dutch novels, van

Cranenburgh et al. (2019) concluded 'that the (literary) quality ratings are inherently difficult to predict from textual features'. We would like to add to this statement that it is unlikely that human literariness ratings can be well predicted on the basis of a single textual feature like stepwise distance or any other for that matter. Thus, Jacobs and Kinder (2019) predicted the rated literariness of >400 poetic and nonpoetic metaphors using an initial set of >70 features. Their findings suggest a set of 11 features that could influence the "literariness" of metaphors, including their sonority score, length and surprisal value.

**Figure 7 here.**

**Figure 8 here**

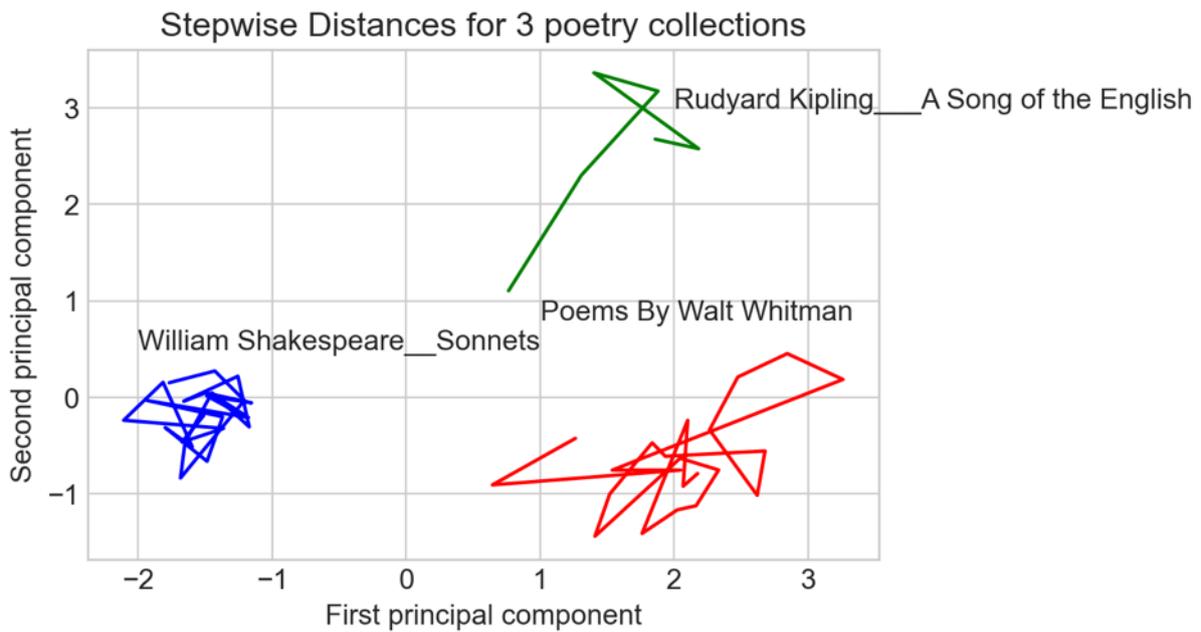

Thus, in summary, as could be expected plays and poetry dominate the hypothetical literariness scale proposed by van Cranenburgh et al. (2019), while stories and essays form the lower end of it with novels and children's literature in the middle. To establish the validity of this scale –which seems to have reasonable face validity and which provided decent predictions of human ratings for a large sample of Dutch novels– definitely more empirical studies of literature are needed. These should also look at differences between authors such as those represented in Figure 8.

**4.3 Forward Flow**

While the above semantic complexity measures likely reveal *one* aspect of the literariness of texts, another recent computational measure –based on latent semantic analysis– and termed *forward flow* is supposed to reflect the *creativity* of writing (Gray et al., 2019; http://www.forwardflow.org/home). The forward flow measure (FF) is similar to the SWD measure, but at the level of single words: for a single word in a sequence it is calculated by the average semantic distance of that word from all preceding words. For the entire sequence FF is just the average of the FFs for all words (cf. equ. 4 in Table 2).

It should be noted that the LSA models used by the creativity (FF) engine were built from the same Touchstone Applied Sciences Associates (TASA) corpus used by Landauer and colleagues on their LSA server (http://lsa.colorado.edu/). For analyzing the language of the poetry of Shakespeare this is not the most adequate corpus and model. Therefore, here we used the present

GLEC corpus which includes all 154 Shakespeare sonnets to generate a DSM (cf. Jacobs, 2018a,b).

We first ran an ANOVA with the factor text category (six levels) on the FF measure and established both a significant overall effect (FF: $R^2$ = .36, F ratio = 303, p <.0001) and significant differences between categories. The obtained FF rank order was: poems > plays > essays >= stories >= children > novels ('>' means a significant difference at the p < .0001 level), '>=' a non-significant one). The two 'most creative' categories match those of the preceding 'literariness' indices (cf. 4.2), albeit in reversed order. However, for some reason novels come last on the present FF scale, while they were 3[rd] for both literariness indices. This somehow counterintuitive finding requires closer inspection in future studies which should also use empirical cross-validation with human rating data.

While such a study goes beyond the aims and scope of this paper, we took a closer look at differences between the FF scores for the 112 authors across all text categories, summarized in Table 5 for 20 authors. Based on the FF index, quite plausibly the leading 'most creative' authors in GLEC all are poets: Milton, Pope, Keats, Byron, Wordsworth, Swift, and Shakespeare. On the low end of the FF scores, we find a mix of essayists and novelists: Lincoln, Stockton, Pyle, Trollope, Defoe, Bunyan, and Newton. We refrain from interpreting these results and leave this task to literary scholars, our main aim being to propose potentially relevant quantitative predictors of semantic complexity and literary quality or attractiveness which can serve in future theoretical or empirical studies of literature.

**Table 5. Top and flop 10 authors according to the FF measure.**

|     | Author | Number of texts | Mean FF |
| --- | --- | --- | --- |
| 1.  | John Milton | 8 | 0,878 |
| 2.  | Alexander Pope | 5 | 0,8762 |
| 3.  | John Keats | 5 | 0,8754 |
| 4.  | Lord Byron | 5 | 0,872 |
| 5.  | William Wordsworth | 5 | 0,8708 |
| 6.  | Jonathan Swift | 5 | 0,8702 |
| 7.  | William Shakespeare | 190 | 0,869478947 |
| 8.  | Robert Browning | 5 | 0,8656 |
| 9.  | Philip Kindred Dick | 11 | 0,863 |
| 10. | Elizabeth Barrett | 6 | 0,859833333 |

|  | Browning | | |
|---|---|---|---|
|  | ----------------------- | | |
| 11. | James Otis | 27 | 0,828814815 |
| 12. | William Henry Hudson | 12 | 0,826333333 |
| 13. | William Dean Howells | 65 | 0,826076923 |
| 14. | Abraham Lincoln | 16 | 0,8245 |
| 15. | Frank Richard Stockton | 31 | 0,823354839 |
| 16. | Howard Pyle | 5 | 0,8228 |
| 17. | Anthony Trollope | 71 | 0,822492958 |
| 18. | Daniel Defoe | 13 | 0,819384615 |
| 19. | John Bunyan | 8 | 0,81575 |
| 20. | Sir Isaac Newton | 5 | 0,7888 |

## 4.4 Kolmogorov Complexity and Book Beauty: a 1<sup>st</sup> exploration

The preceding sections showed considerable variance in the semantic complexity measures used to estimate the literariness and creativity of texts. In this final section of part 2 we complement these measures by a novel measure hypothetically predicting the *perceived beauty* of a text. Evaluating a story or book as more literary than another does not necessarily predetermine this text as being perceived as more beautiful or likeable. Semantic deviation from a norm by itself is not sufficient to predict perceived beauty (e.g., Kintsch, 2012; Schmidhuber, 1997). Thus, these latter authors proposed a novel complexity measure, Kolmogorov complexity (cf. equ. 5 in Table 2), as an index of a text's perceived beauty hypothesizing that beauty in a certain type of art can be identified with low Kolmogorov complexity. The core idea behind Kintsch's (2012) simplified model of beauty for textual art is that the mental representation of a text follows the principle of complexity reduction. If this mental representation of a book is estimated via topic analysis (cf. 3.1) one can easily compute two measures, explained in the next section, that will contribute to the perceived beauty of that book.

**Harmony**. Kintsch proposed that a story's or book's harmony can be computed as the average distance between the topic probability distribution of the entire book and those of its chapters (cf. equ. 6 in Table 2). A harmonious book requires that it is possible to encode the whole book as well as its parts (i.e., chapters) with *the same mental model*, that is, the cost of transforming its chapters $C_1, C_2, \ldots, C_n$ into the whole must be small, i.e. minimizing Kolmogorov complexity. As a practical and metrical substitute for the asymetric (non-metrical) Kullback-Leibler divergence originally proposed by Kintsch (2012) for measuring harmony, here we used the symmetrical

Hellinger-Bhattacharyya distance (HBD in the following) included in the *gensim* topics analysis library.

**Variety**. According to Kintsch (2012) harmony is a necessary but not a sufficient condition for artful beauty, because very boring objects could be generated by solely minimizing Kolmogorov complexity. Therefore, a 2nd requirement for a harmonious object to be perceived as beautiful is that its aspects or parts are varied. All parts must fit the whole, but the parts must be distinct from each other. Thus, variety (cf. equ. 7 in Table 2) requires that the cost of transforming one aspect into another is large, or, more particularly larger than the cost of transforming it into the whole. With regard to a 'beautiful' book this means that the HBD between two (or more) chapters should be greater than that between each chapter and the book as a whole.

To sum up, if two books have about equally large values of harmony, but one has a notably greater value of variety than the other its theoretical beauty would also be greater, and vice versa. As a tentative simplified measure of book beauty[2] we computed the harmony and variety for all texts in GLEC that included information about chapters in four of the six categories, i.e. children, essays, novels and stories. Regarding harmony, the rank order of the mean values was (see Table X): essays (.098) > novels (.092) >= stories (.091) >= children/CYL (.088). For variety, the rank order of the mean values was slightly different: novels (.087) >= essays (.086) >= stories (.084) >= children (.083). However, the only statistically significant differences in this rank order were between novels and essays on the one hand, and children on the other.

What makes the interpretation of these data more difficult than for the preceding measures is the fact that theoretically there is no simple linear relationship between the index and its two measures, i.e. beauty does not simply increase when both harmony and variety become greater. Rather there is an optimum pairing, since according to equ. (7), the semantic distance between two or more chapters (i.e., VARI) must be greater than that between each chapter and the whole book (i.e., HARM). The mean values (averaged across all texts in each category) in Table X do not vary enough to allow an interpretation, especially since harmony and variety were highly linearly correlated in our sample (e.g., r = .87 for novels, p < .0001).

---

[2] Kintsch (2012) also proposed a third measure in his model, called compression or dynamics, referring to the principle of increasing harmony over time: 'A beautiful object allows for the compression of complexity; it affords the discovery of new harmonies'. With regard to books, this 3rd measure requires that one discovers new topics in a book and its chapters that reduce Kolmogorov complexity with regard to the results of a previous topic analysis. However, Kintsch never implemented or tested his model stating that 'A serious test of such a model would be a major undertaking, requiring a careful selection of representative texts.' As the 1st actual implementation of this model, here we opted for simplifying things to the first two (static) measures which theoretically reflect the 1st reading of a book. An empirical validation of the 3rd (dynamic) measure would require the re-reading of books which is still a big challenge for current studies in neurocognitive poetics.

However, the example in Figure 9 helps clarify the concept. To illustrate how the theoretical beauty of books may vary consider the six novels of Jane Austen. According to Kintsch's (2012) criteria, 'Emma' would have the highest beauty potential of all six, since it combines the 2$^{nd}$ smallest harmony value (smaller values = greater harmony) with the 3$^{rd}$ highest variety value and fullfills equ. 7, i.e. 0.103 (VARI) > 0.097 (HARM; cf. Kintsch's, 2012, hypothetical example in Table 1). In comparison, 'Pride and Prejudice' has greater harmony (0.091), but also smaller variety (0.099). Whether this purely theoretical prediction meets expert evaluations or naïve readers' subjective feelings is an intriguing open question and, of course, such a simplified two-feature model will not capture the full beauty potential of artful texts. Still, this 1$^{st}$ exploratory quantification of Kintsch's (2012) conceptual model provides enough predictive power for future validation studies and the basis for more sophisticated competitive models to come.

**Figure 9 here**

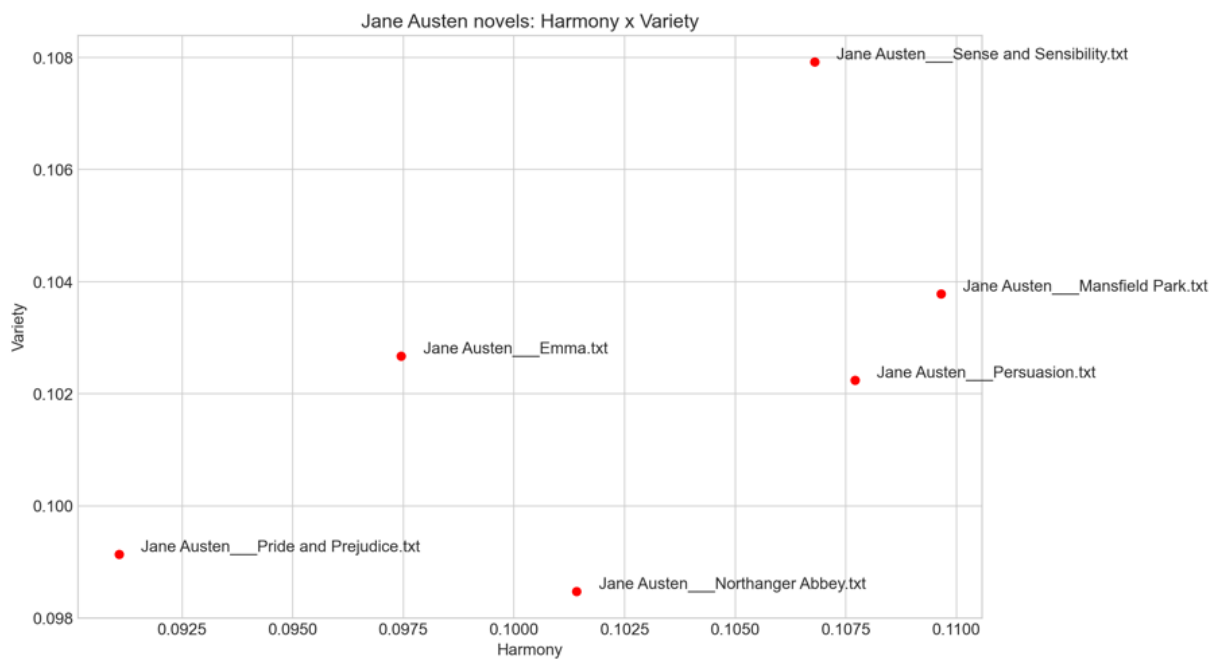

## 5 Experiments on Text classification and Authorship Recognition

The above studies providing quantitative analyses have shown that the GLEC is a rich corpus offering sufficient variation for scientific studies of literature on a number of dimensions such as topics, sentiment, or semantic complexity. For lack of empirical data providing ground truths, these descriptive studies could not test any predictions beyond certain face validities, e.g. that poetry offers higher semantic complexity and creativity than CYL. In part 5 of this paper we therefore tested the predictive validity of five measures of semantic complexity in text classification and authorship recognition experiments.

**Text classification and category recognition**

Text classification is a standard task in machine learning assisted NLP usually tackling the problem of correctly labeling test texts from newspapers or internet sources into genre categories like business or politics, or literary texts into categories like adventure or romance. An early influential NLP study[18] of the ~1 million word *Brown corpus*[2] used easy-to-compute style features requiring no tagging (e.g., sentence length, TTR) fed into a simple feedforward neural net (a Multi-Layer Perceptron/MLP) and achieved maximum accuracies of about 80% for *binary* classifications such as whether a text is a narrative or not. More recent work on literary texts sampled from the Gutenberg project introduced affective semantic content features like joy or anger for genre classification (e.g., western or romance) fed into a decision tree algorithm which achieved a maximum *multiclass* classification accuracy of ~80% for the category science fiction (Samothrakis & Fasli, 2015)[4]. Using a combination of five surface and five affective semantic features, we could recently show that quasi-error free text classification and authorship recognition is possible for the GLEC texts (Jacobs & Kinder, 2020). However, as far as we know, the above measures of semantic complexity have not yet been tested for their usefulness in text classification and authorship recognition.

**Machine learning**

The classification tasks were all performed using the *Predictive Modeling Platform* of JMP 15 Pro successfully employed in previous papers (Jacobs et al., 2017; Xue et al., 2019..)[10,11,14,33]. A standard classifier from the platform was used: Neural Networks/MLP. The MLP had the same hyperparameters as in Jacobs and Kinder (2020): one hidden layer with 100 units using the nonlinear hyperbolic tangent (*TanH*) activation function, and a 2nd hidden layer with 25 units using the linear identity (Lin) activation function. The tours parameter was set to 10 iterations for each task, the penalty parameter was set to square. The cross-validation procedure was k-fold with k = 5. The random seed parameter was always set to 1 ensuring that the present results can be exactly reproduced. We renounced on hyperparamter tuning, since we were not interested in any 'benchmark boosting' (cf. Jacobs & Kinder, 2020), but wanted to check the utility of the present five semantic features for standard NLP classification tasks.

Since the harmony and variety measures could not be computed for poems and plays, text classification was run on the remaining four categories. The training sets contained 1010 texts, the test sets 253. Figure 10 shows the ROC curve indicating an excellent performance: overall misclassification rate was 7.5%; the AUC values were all above .98. The confusion matrix in Table 6 shows that novels were recognized best followed by essays, children's literature and stories which were confused with both children's literature and essays in 12% of the cases. The overall feature importances[3] were the following: SWD: .45, FF: .28, ITV: .278, VARI: .175, and HARM: .172. They show a clear winner, SWD, as the main contributor to this excellent text classification performance.

**Table 6. Classification accuracy (confusion matrix) and Overall Feature Importances (MLP model) for the four categories (validation set).**

| tags | children | essays | novels | stories | SWD | ITV | FF | HARM | VARI |
|---|---|---|---|---|---|---|---|---|---|
| **children** | 0,875 | 0,031 | 0,062 | 0,031 | | | | | |
| **essays** | 0 | 0,944 | 0 | 0,055 | | | | | |
| **novels** | 0,023 | 0 | 0,976 | 0 | | | | | |
| **stories** | 0,12 | 0,12 | 0 | 0,76 | | | | | |

**Figure 10 here**

---

[3] In the current study, feature importances were computed as the total effect of each predictor as assessed by the *dependent resampled inputs* option of JMP14 Pro. The total effect is an index quantified by sensitivity analysis, reflecting the relative contribution of a feature both alone and together with other features. This measure is interpreted as an ordinal value on a scale of 0 to 1, values > 0.1 being considered as important (Strobl et al., 2009).

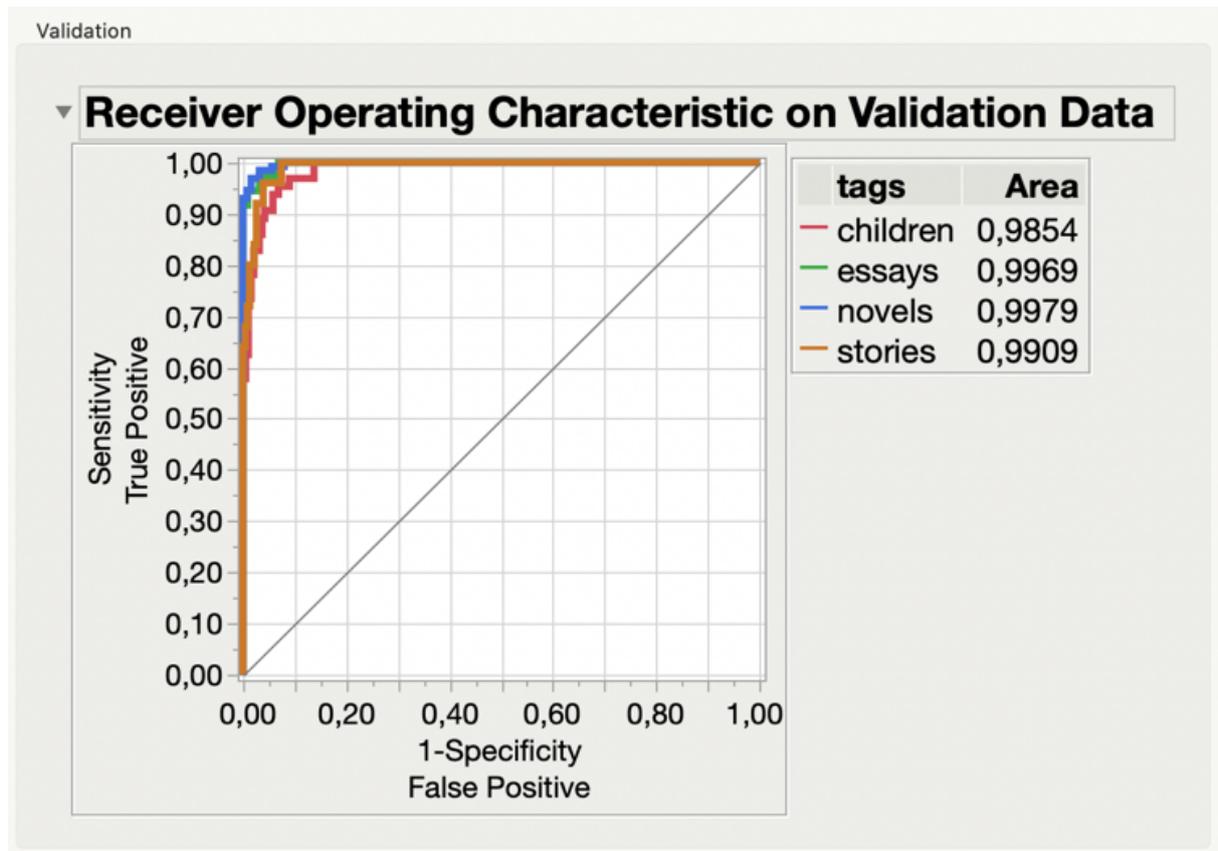

**Category Footprints**

The MLP computes both overall feature importance values (Table 7) and specific ones for each text category. These can be used to visualize a sort of category footprint or profile, as shown in Figure 11 (Y axis = feature importances). Complementing the overall feature importance data from Table 7, such profiles are useful for identifying which mix of literariness (ITV, SWD), creativity (FF), and beauty (HARM, VARI) indices best characterises each category and for finer-grained comparisons between different subcorpora. The data show similarly shaped curves for all four categories for which we could compute five features with a peak on the SWD feature and a dip for HARM. Thus, differences in amplitude (i.e., feature importances) rather than in shape (rank order of features) revealed what category a GLEC text came from, although for 2/4 ITV was the 2[nd] most important feature and FF for the other two. The data also show that, on its own, the SWD feature appears to be diagnostic in discriminating between stories and essays, but is of little help for telling the other two categories apart. Also, to discriminate stories from children's books, FF is not diagnostic. In future studies the present category footprints could be compared with those of other corpora to check which features are *generally* useful and which are more *specific* to certain corpora of a certain time period and with a certain genre and author collection.

**Figure 11 here**

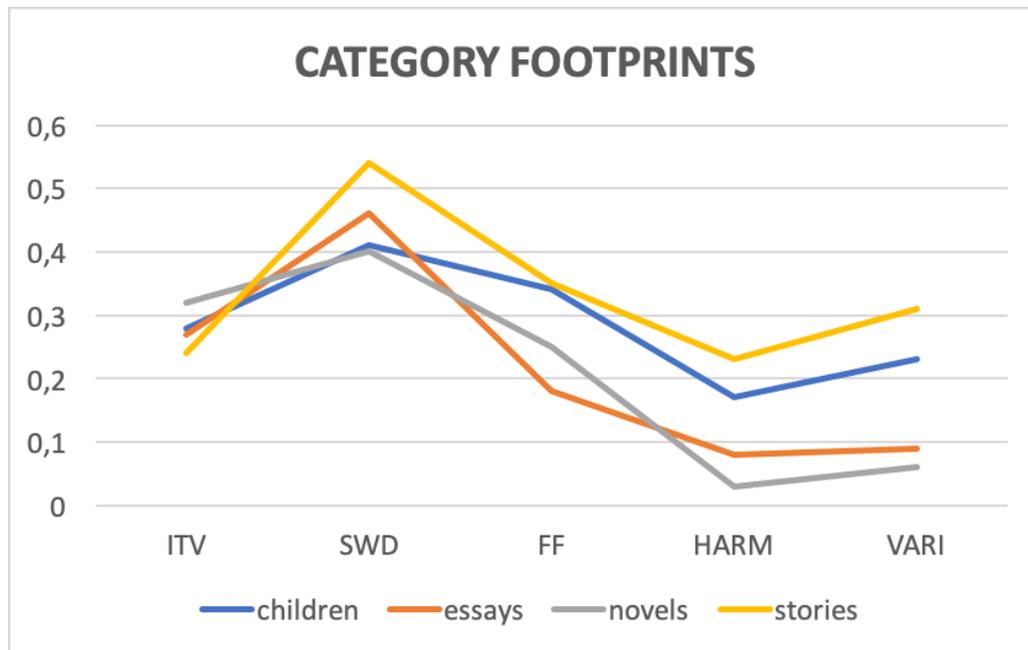

**Authorship recognition**

The second experiment concerned authorship recognition (attribution), a pivotal task in NLP ever since Mosteller and Wallaces' (1964) classical study. Despite considerable progress in stylometrics and ML techniques it remains a challenge in computational research on literature focusing on issues such as the choice of optimal features and classifiers (Jockers & Witten, 2010), or minimal sample length for obtaining stable results (Eder, 2015). Here we tested i) how well the MLP performed in recognizing the altogether 112 different authors of the present GLEC books, ii) whether success rate depended on text category, and iii) which of the semantic features were the most diagnostic.

**Table 7. Confusion Matrix (%) and Feature Importances (MLP model) Author Recognition**

| text category | $R^2_{ent}$ | % misclass. | SWD | ITV | FF | VARI | HARM |
|---|---|---|---|---|---|---|---|
| **CYL** | .95 | .016 | **.43** | .41 | .39 | .27 | .23 |
| **ESS** | .91 | .027 | **.53** | **.53** | .5 | .5 | .44 |
| **NOV** | .79 | .28 | **.47** | .43 | **.47** | .32 | .29 |
| **PLA** | .86 | .04 | .35 | **.55** | .51 | - | - |
| **POE** | .74 | .20 | .56 | **.65** | .64 | - | - |
| **STO** | .97 | 0 | .46 | .47 | **.59** | .49 | .58 |

The data in Table 7 first show that overall, as could be expected for the present sample, authorship recognition is more difficult than text classification with misclassification rates between 0 and 28%. Second, there are large differences in performance as a function of text category: authors of stories were easy to recognize, authors of novels much less so. Third, given that only five and three features were used here, respectively, their performance in this difficult task is quite promising[4]. The two indices of literariness (SWD, ITV) were most diagnostic in 5/6 cases, the creativity index FF twice, and the two beauty indices –which participated in only four categories–, both came second once: HARM in recognizing authors of stories and VARI for essays.

**Author Footprints**

As for text categories, here we looked at author footprints to see which features characterise certain authors best. A few illustrative examples from essays and novels are given in Figure 12. Regarding essays, SWD and HARM seem to characterise Lewis Carrolls writings in GLEC best, the latter clearly being Oscar Wilde's core feature in this set of five –in stark contrast to Winston Churchill whose essays are mainly characterised by their creativity index (FF). Figure 12b (right) allows comparing the same author's distinctive features across two different text categories by showing that Winston Churchill has quite different profiles for his essays and his novels, the latter being mainly characterised by the literariness indices ITV and SWD. For recognizing Jane Austen's novels a literariness feature (ITV) and a beauty feature (VARI) are most diagnostic, for PG Wodehouse it is HARM and FF. Thus, an interesting mix of literariness, creativity, and beauty indices characterises each author in GLEC. In future studies it will be interesting to extend this set of five semantic features by style (surface) features such as TTR and other content features such as affective-aesthetic potential (cf. Jacobs & Kinder, 2020) to reveal a particular author's overall aesthetic style and emotion potential.

---

[4] As mentioned before, no hyperparameter tuning was used to improve performance.

**Figure 12 here**

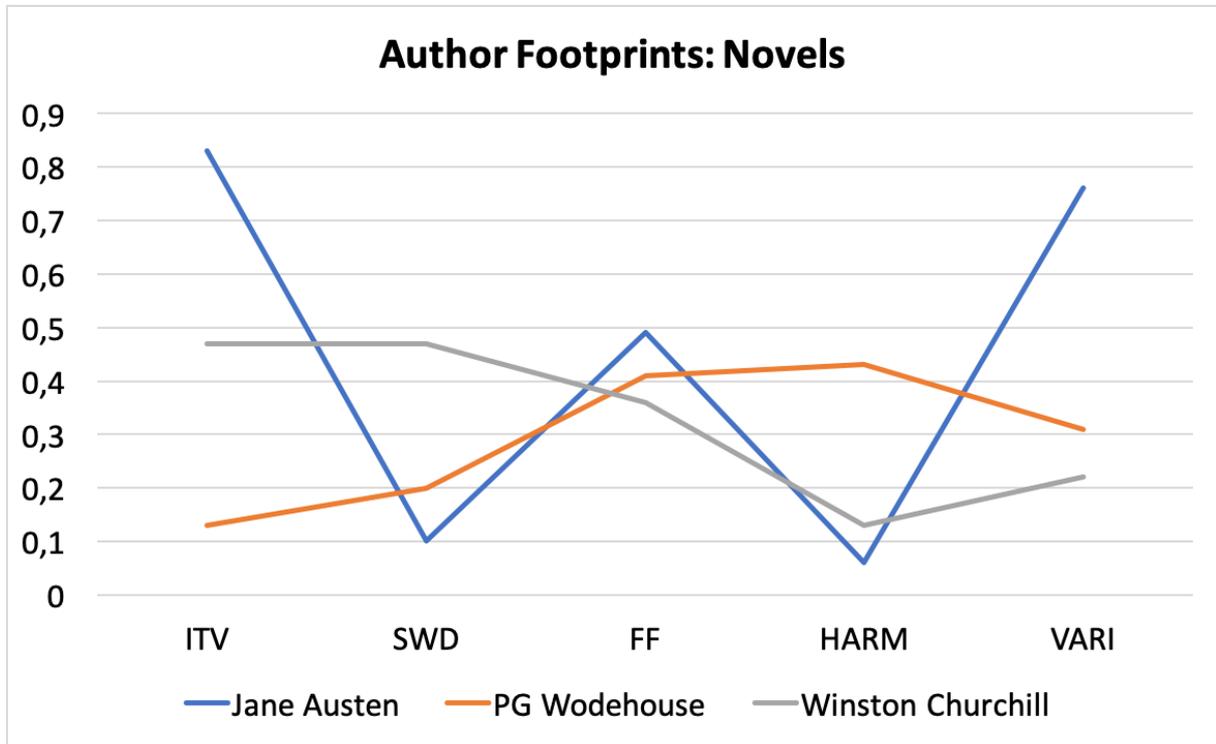

To summarize, the two experiments on standard NLP tasks yielded promising results regarding the utility of the five semantic complexity measures for future applications in computational and empirical studies of literature. Each of the five features was diagnostic in both text classification and authorship recognition (as indicated of feature importance values well above .1; cf. Strobl et al., 2009) with the two literariness indices proposed by van Cranenburgh et al. (2019) somewhat standing out. Combining these features with previously used sets of surface and affective semantic features (e.g., the 10 features in Jacobs & Kinder, 2020) and applying them to other corpora than GLEC would be a valuable task for determining an optimal or minimal 'standard set' of features for future benchmarks in text category and authorship recognition.

**6 Discussion and Outlook**

The results of three studies revealed interesting differences in the main topics and sentiments between both text categories and authors. They also showed that the phenomenon of *vocabulary decay* is general to all six categories of GLEC and suggest that prolific writing by itself (e.g., >20 works in GLEC) does not increase its likelihood, but that time pressure (e.g., Charles Dickens vs. Jane Austen) is a crucial factor (Elliott, 2016). Moreover, computation of several measures of semantic complexity revealed that plays are the most literary texts in GLEC, followed by poems and novels, and that poems and plays are the most creative categories. In a first exploratory study, a novel index of the perceived beauty of verbal art predicted that 'Emma' is the theoretically most beautiful of Austen's novels. Finally, the 3$^{rd}$ study demonstrated that these novel measures of semantic complexity are important features for text classification and authorship recognition with overall predictive accuracies in the range of .75 to .97. Our data pave the way for future computational and empirical studies of literature or experiments in reading psychology and offer multiple baselines and benchmarks for analysing and validating other book corpora.

Thus the present approach could serve as a model applicable to other corpora in different languages or from different time periods, such as the German *childLex* corpus (Schroeder et al., 2015). It could be used, for instance, for testing the assumption that children and adolescents of different age groups differentially develop sensitivities for text features like 'happiness', likely depending on their overall reading experience, literary background and personal preferences. A recent literature review on the possible causes of liking verbal materials ranging from single words to entire poems and books (Jacobs et al., 2016) suggests that a main driving force behind human liking ratings is the extent to which such high-dimensional stimuli are associated with basic emotions such as joy/happiness or disgust.

Regarding the various semantic complexity measures, the present results are promising for future computational and empirical validation studies. Which of these indices proves to be optimal for estimating the literariness, creativity or beauty of literary texts and which alternative indices may successfully complement them indeed seem central issues deserving empirical investigation.

**Figure Captions**

**Figure 1a-f.** Top topics for each of the six text categories (alphabetical order from top to bottom: CYL, ESS, NOV, PLA, POE, STO) in the form of word clouds.

**Figure 2.** 'happiness' and 'fear' scores as computed by *SentiArt* for the six text categories.

**Figure 3a-f.** 'happiness' and 'fear' scores as computed by *SentiArt* for the all authors in the six text categories.

**Figure 4a-f.** Top topics for selected authors: a) Alcott (top row, left), b) Bunyan (top row right), c) Barrie (middle row, left), d) Defoe (middle row, right), e) Kipling (bottom row, left) and f) Keats (bottom row, right).

**Figure 5a-c.** Entropy (H) as a function of text segment (1-100) for selected authors.
**Figure 6**. Entropy (H) as a function of text segment for the six text categories.

**Figure 7.** ITV and SWD measures for authors in the six text categories.

**Figure 8.** SWD ($1^{st}$ and $2^{nd}$ principal components) for three selected authors of poetry.
**Figure 9**. Harmony and Variety for Jane Austen's six novels.
**Figure 10.** Receiver Operating Characteristic (ROC) curves for the four text categories (validation set).
**Figure 11.** Category Footprints based on Feature Importances/FI
**Figure 12.** Author Footprints based on FI values (probability that a given feature is important for authorship recognition) for three exemplary authors from the POE category.

**Acknowledgements**

None

**Author contributions**

AJ designed and performed the study. AK and AJ performed the ML part and analysed the results. AJ drafted the manuscript, AJ and AK contributed both to the final version of the paper.

**Competing interests**

The authors declare no competing interests.

**Additional information**

**Data availability.** For reproducing the statistical/machine learning analyses of the main data table, the JMP14 Pro software (not free) is necessary. A .xlxs version of the main table will be made publicly available, though, at github or alternative outlets for users who wish to analyze the data with alternative software. A table listing the 2722 texts and the texts themselves will also be made publicly available and can be obtained by the authors on demand, as can be the original log files from JPM14 Pro for the machine learning analyses (in .pdf format).

**Code availability.** All custom code will be made publicly available at github or alternative outlets (e.g., a special server of FU Berlin).

Correspondence and requests for materials should be addressed to AJ.